\documentclass[letterpaper,journal]{IEEEtran}
\usepackage{cite}
\usepackage{amsmath,amssymb,amsfonts}
\usepackage{algorithmic}
\usepackage{graphicx}
\usepackage{textcomp}
\usepackage{xcolor}
\usepackage[caption=false,font=normalsize,labelfont=sf,textfont=sf]{subfig}
\usepackage{tabularx}
\usepackage[table]{xcolor}
\usepackage{booktabs}
\usepackage{multirow}
\usepackage{bm}
\usepackage{verbatim}
\usepackage{hyperref}
\usepackage{newtxtext,newtxmath}

\def\BibTeX{{\rm B\kern-.05em{\sc i\kern-.025em b}\kern-.08em
    T\kern-.1667em\lower.7ex\hbox{E}\kern-.125emX}}
\begin{document}

\title{Multimodal Optimal Transport for Training-free Temporal Segmentation in Surgical Robotics
}

\author{
\IEEEauthorblockN{
Omar Mohamed$^1$, Edoardo Fazzari$^{1*}$, Ayah Al-Naji$^1$, Hamdan Alhadhrami$^1$, Khalfan Hableel$^1$, Saif Alkindi$^1$, Ivan Laptev$^1$, Cesare Stefanini$^1$\thanks{$^1$Authos are with the Dept. of Robotics, Mohamed bin Zayed University of AI, Abu Dhabi, UAE. \{\texttt{omar.mohamed}, \texttt{edoardo.fazzari}, \texttt{ayah.al-naji}, \texttt{hamdan.alhadhrami}, \texttt{khalfan.hableel}, \texttt{saif.alkindi},  \texttt{ivan.laptev}, \texttt{cesare.stefanini}\}\texttt{@mbzuai.ac.ae}}
}
}

\maketitle

\begin{abstract}
Automated recognition of surgical phases and steps is a fundamental capability for intraoperative decision support, workflow automation, and skill assessment in robotic-assisted surgery. Existing approaches either depend on large-scale annotated surgical datasets or require expensive domain-specific pretraining on thousands of labeled videos, limiting their practical deployability across diverse robotic platforms and clinical environments. In this work, we propose TASOT (Text-Augmented Action Segmentation Optimal Transport), an annotation-free framework for surgical temporal segmentation that requires no task-specific annotations or surgical-domain pretraining. TASOT extends the Action Segmentation Optimal Transport (ASOT) formulation by incorporating temporally aligned textual descriptions generated directly from the input video, fusing visual and semantic cues within a unified unbalanced Gromov-Wasserstein optimal transport objective. Visual representations are extracted using DINOv3, while temporal captions produced by a vision-language model are encoded via CLIP and temporally aligned to individual frames, providing complementary semantic structure to the transport cost. We evaluate TASOT on three public surgical datasets and four benchmark settings spanning laparoscopic and robotic procedures, showing substantial improvements over the strongest zero-shot baselines: +18.9 F1 on Cholec80, +33.7 on AutoLaparo, +23.7 on StrasByPass70, and +4.5 on BernByPass70. These results suggest that fine-grained surgical workflow understanding in robotic settings can be achieved without manual training annotations or surgical-specific pretraining pipelines, offering a promising alternative for real-world robotic surgical systems.
\end{abstract}


\section{Introduction}
Understanding surgical workflow at the temporal level is a 
fundamental capability for intelligent robotic-assisted surgery. 
Automated recognition of surgical phases and steps enables 
intraoperative guidance, real-time decision support, automated 
skill assessment, and the foundation for higher-level surgical 
autonomy~\cite{RL2026survey, HVQ2025unsupervised}. 
As robotic surgical platforms generate increasingly large volumes 
of procedural video, the ability to parse and understand these 
workflows without manual intervention has become a critical 
bottleneck in the development of clinically deployable surgical AI.

Surgical video, however, presents unique perceptual challenges 
for automated systems. The visual scene is highly complex and 
dynamic, with frequent occlusions, camera motion, instrument 
motion, tissue deformation, and ambiguous anatomical 
structures~\cite{cholec80-2017dataset}. These factors make it 
difficult to distinguish actions based solely on visual 
appearance~\cite{fazzari2025artemis,MIL-NCE2020end}. Moreover, 
approaches that rely on fully supervised segmentation typically 
require dense annotations for surgical videos, which are 
extremely expensive, as each frame must often be labeled by a 
medical expert. Collecting large annotated surgical datasets is 
therefore laborious and time-consuming~\cite{hashemi2023acquisition}.

To reduce the annotation burden, most recent efforts have 
explored “zero-shot” or transfer-based approaches in the 
surgical domain~\cite{SurgVLP2025ZeroShot}. Although these 
approaches achieve strong performance without requiring dense 
annotations, they typically still rely on large pretrained 
networks and employ general, complex architectures that are not 
specifically designed to enforce temporal segment 
structure~\cite{CLIP2021ZeroShot}. In short, existing surgical 
video segmentation methods either demand heavy annotation effort 
or depend on complex pretrained models, and they do not explicitly exploit the structure of OT-based temporal segmentation. Given these limitations, we ask the following 
question: \emph{Is large-scale surgical pretraining truly 
necessary for effective temporal segmentation, or can an 
annotation-free approach achieve competitive performance?}

To address this question, we propose TASOT 
(Text-Augmented Action Segmentation Optimal Transport), an 
annotation-free method for surgical phase and step recognition 
that extends Action Segmentation Optimal Transport 
(ASOT)~\cite{ASOT2024unsupervised} by incorporating textual 
information generated directly from videos. Crucially, TASOT 
requires no surgical-domain annotations, no manual 
task-specific labels, and no narrated footage. Instead, we use off-the-shelf encoders for video frames and associated textual data, and let an optimal 
transport (OT) formulation fuse them. In other words, TASOT 
aligns video and text sequences within a unified framework, 
where a temporally consistent unbalanced Gromov--Wasserstein 
OT~\cite{titouan2019sliced} formulation ensures coherent 
segment boundaries across both modalities. 

We evaluate TASOT on three publicly available datasets, i.e., 
Cholec80~\cite{cholec80-2017dataset}, 
AutoLaparo~\cite{AutoLapro2022dataset}, and 
MultiBypass140~\cite{MultiByPass2024dataset} from both centers. 
Across all datasets, TASOT consistently outperforms existing 
zero-shot models in terms of F1 score, demonstrating that 
surgical temporal understanding can be achieved without 
large-scale surgical pretraining.

In summary, our contributions are:
\begin{itemize}
    \item[$\bullet$] We propose the first multimodal OT-based 
    framework for annotation-free temporal segmentation in the 
    surgical domain, introducing a formulation that integrates 
    visual and textual cues within a unified optimal transport 
    objective, regularized by temporally consistent 
    Gromov--Wasserstein constraints.
    \item[$\bullet$] We demonstrate strong performance in 
    annotation-free surgical temporal segmentation, consistently 
    outperforming existing zero-shot methods on F1 scores across 
    multiple benchmark datasets.
\end{itemize}

\section{Related Work}

Surgical workflow recognition has progressed from fully 
supervised CNN-based architectures~\cite{cholec80-2017dataset} 
through recurrent and temporal convolutional 
networks~\cite{TeCNO2020supervised, MTMS-TCN2021supervised} 
to transformer-based models capturing long-range temporal 
dependencies~\cite{LoViT2025supervised}. More recently, 
self-supervised and adaptive transformer-based approaches 
have also been explored for robotic surgical step recognition, 
improving robustness across surgeons, centers, and imaging 
conditions~\cite{ye2026self}. While these 
approaches achieve strong performance, they require dense 
frame-level annotations from clinical experts for downstream 
workflow recognition---a prohibitive bottleneck for deployment 
across diverse robotic platforms and clinical 
environments~\cite{hashemi2023acquisition, MultiByPass2024dataset}.

To reduce annotation burden, video-language pretraining methods 
have emerged for zero-shot surgical recognition. Building on 
contrastive frameworks such as CLIP~\cite{CLIP2021ZeroShot} and 
MIL-NCE~\cite{MIL-NCE2020end}, domain-specific adaptations 
including SurgVLP~\cite{SurgVLP2025ZeroShot}, 
HecVL~\cite{HecVL2024ZeroShot}, and 
PeskaVLP~\cite{PeskaVLP2024ZeroShot} align surgical video with 
narrated transcripts via contrastive and hierarchical objectives, 
achieving competitive zero-shot performance. However, since 
paired surgical video-text corpora are scarce, these 
methods often rely on multi-stage text acquisition pipelines: 
SurgVLP and HecVL use Automatic Speech Recognition (ASR) 
systems to transcribe narrated surgical lecture videos, while 
PeskaVLP further processes these transcripts using 
GPT-4~\cite{achiam2023gpt} to produce structured 
procedural summaries. These pipelines are therefore most 
naturally applicable to procedures where large-scale narrated 
surgical footage is available---a condition less common in 
standard robotic operating room settings where surgeries are 
performed without narration. Furthermore, the resulting models 
depend on pretrained multimodal backbones learned from curated 
video-text pairs. In contrast, TASOT generates semantic 
descriptions directly from silent input video at inference time 
using a vision-language model, requiring neither narrated 
surgical footage, ASR transcription, nor offline corpus 
construction.

In parallel, unsupervised temporal action segmentation aims to 
partition untrimmed videos into coherent segments without 
frame-level annotations~\cite{temporal2024survey}. Early 
approaches learn frame embeddings and cluster them under 
temporal regularization~\cite{TW-FINCH2021unsupervised,
sener2018unsupervised}, later refined through 
reconstruction~\cite{VTE2021unsupervised} or discriminative 
embedding objectives~\cite{UDE2021unsupervised}. More recently, 
Optimal Transport (OT) has provided a principled framework for 
joint representation learning and pseudo-label assignment. In 
particular, ASOT~\cite{ASOT2024unsupervised} formulates 
segmentation as an unbalanced Gromov--Wasserstein OT problem 
without predefined action ordering, achieving strong 
general-domain performance. HVQ~\cite{HVQ2025unsupervised} 
further improves short-action detection via hierarchical vector 
quantization, but its codebook-based formulation does not 
explicitly couple representation learning with clustering as in 
OT-based alignment. Nevertheless, existing OT-based approaches 
operate purely on visual features. To date, multimodal textual 
cues have not yet been integrated into an OT framework for 
surgical temporal segmentation.

\section{Method}

\begin{figure*}[!t]
    \centering
    \includegraphics[width=1\linewidth]{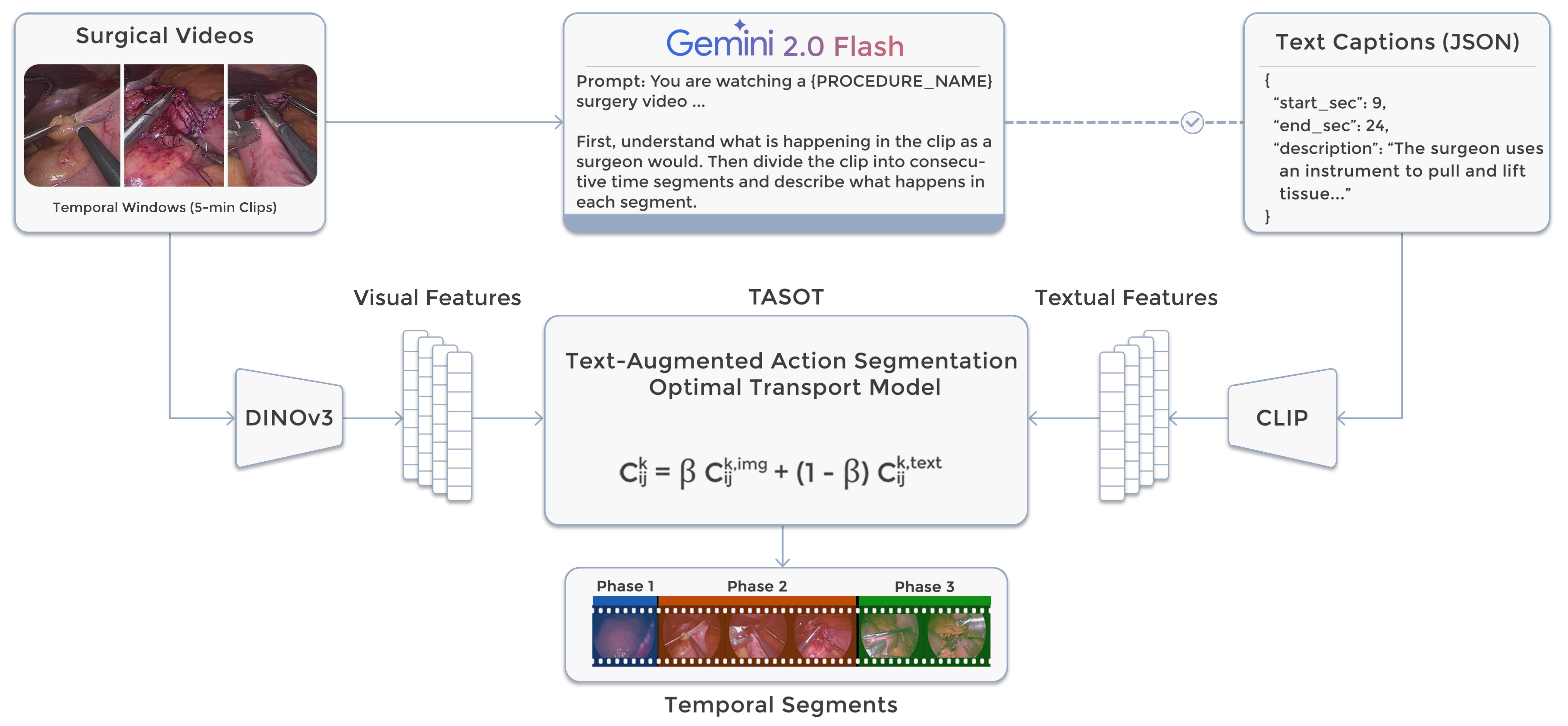}
    \caption{\textbf{Overview of TASOT.} Surgical videos are divided into temporal windows and processed by a vision-language model to generate structured temporal captions. Visual features extracted with DINOv3 and temporally aligned textual features encoded with CLIP are integrated within TASOT through a weighted multimodal transport cost, enabling annotation-free surgical temporal segmentation.}
    \label{fig:TASOT_Framework}
\end{figure*}

We introduce \textbf{TASOT} (Text-Augmented Action Segmentation 
Optimal Transport), an annotation-free multimodal framework for 
surgical temporal segmentation that extends 
ASOT~\cite{ASOT2024unsupervised} by incorporating temporally 
aligned textual descriptions generated directly from the input 
video. As illustrated in Fig.~\ref{fig:TASOT_Framework}, TASOT 
operates in three stages: (i) a captioning pipeline that 
processes raw surgical video through a vision-language model 
to produce temporally grounded natural language descriptions; 
(ii) a feature extraction stage that encodes video frames 
using DINOv3~\cite{DINOv32025encoder} and caption segments using 
CLIP~\cite{CLIP2021ZeroShot}, producing temporally aligned 
visual and textual feature streams; and (iii) a multimodal 
optimal transport stage that integrates both streams within 
a unified unbalanced Gromov-Wasserstein objective to produce 
temporally coherent segment assignments. Crucially, TASOT 
requires no surgical-domain annotations, no manual 
task-specific labels, and no narrated footage.
Instead, semantic information is obtained by generating 
temporally grounded descriptions directly from the input 
video at inference time using a pretrained vision-language 
model. 

\subsection{Captioning Pipeline}

Given a surgical video $v$ of duration $T$ seconds, we first 
divide it into non-overlapping temporal windows of fixed 
length $W$ (default $W = 300$\,s):
\begin{equation}
    w_m = v[s_m, e_m), \quad s_m = mW, \quad 
    e_m = \min\bigl((m+1)W,\, T\bigr),
    \label{eq:windows}
\end{equation}
where $s_m$ and $e_m$ denote the start and end times of 
the $m$-th window. Each window clip is uploaded to Gemini 
2.0 Flash~\cite{GeminiFlash2025}, which processes the raw 
video frames directly and generates a structured temporal 
description of the surgical activity within that clip.

\noindent\textbf{Prompt design.} Gemini is prompted with 
the following template, which was designed to produce 
clinically grounded, reproducible descriptions without 
requiring procedure-specific instructions beyond the 
procedure name:

\begin{quote}
\texttt{You are watching a \{PROCEDURE\_NAME\} surgery 
video. The clip starts at time 0 seconds and lasts 
\{CLIP\_LEN\} seconds. First, understand what is 
happening in the clip as a surgeon would. Then divide 
the clip into consecutive time segments and describe 
what happens in each segment. Return ONLY valid JSON 
with the following fields: start\_sec, end\_sec, 
description. Guidelines: use integer seconds within 
[0, CLIP\_LEN]; write clear descriptions of surgical 
actions mentioning tools and anatomy when visible; 
produce 10--20 segments per 5-minute clip with preferred 
durations of 10--25 seconds; avoid vague phrases such as 
``view change''; use generic tool names only.}
\end{quote}

The model returns a JSON array of consecutive temporal 
segments $\mathcal{S}_m$:
\begin{equation}
\begin{gathered}
\mathcal{S}_m
= \Bigl\{ \bigl(\tau_j^{(m)},\, \tau_{j+1}^{(m)},\, \text{text}_j^{(m)}\bigr) \Bigr\}_{j=1}^{J_m}, \\
0 = \tau_1^{(m)} < \cdots < \tau_{J_m+1}^{(m)} = |w_m|
\end{gathered}
\label{eq:segments}
\end{equation}
where each entry specifies a start time, end time, and 
natural language description of the surgical action 
occurring in that interval. Window-local timestamps are 
converted to global video time as:
\begin{equation}
    t_{j,\text{start}}^{(m)} = s_m + \tau_j^{(m)}, 
    \qquad 
    t_{j,\text{end}}^{(m)} = s_m + \tau_{j+1}^{(m)}.
    \label{eq:global_time}
\end{equation}
Window-level segments are then merged and temporally 
ordered to form the complete video-level caption set.

\noindent\textbf{Robustness and failure handling.} 
The pipeline includes dedicated handling for two 
failure modes. First, if Gemini returns output wrapped 
in Markdown code fences (e.g., \texttt{```json}), 
these are stripped before JSON parsing. Second, if 
JSON parsing fails, for instance due to truncated 
output or timestamp violations, the raw response is 
logged to disk and the window is skipped, allowing the 
pipeline to resume from the last successful window on 
re-execution. Remote files are deleted from the Gemini 
API after each window to avoid storage accumulation. 
In practice, parse failures occurred in fewer than 
$1$\% of windows across our evaluation datasets, 
and re-running the pipeline on failed windows resolved 
all cases.

\subsection{TASOT Framework}

\noindent\textbf{Multimodal Feature Extraction.}
Given a surgical video sampled at 1\,fps, consistent 
with standard surgical workflow annotation 
protocols~\cite{cholec80-2017dataset}, each frame 
$I_t$ is encoded by a frozen DINOv3~\cite{DINOv32025encoder} 
backbone to produce visual frame embeddings 
$x^{\text{img}}_t \in \mathbb{R}^{D}$. DINOv3 is 
selected for its strong spatial representations learned 
through self-supervised pretraining, which transfer 
effectively to the surgical domain without requiring 
any domain-specific fine-tuning.

In parallel, the temporally grounded captions produced 
by the pipeline previously described 
are encoded using CLIP~\cite{CLIP2021ZeroShot}, 
producing segment-level text embeddings $e_j \in 
\mathbb{R}^{D}$. Each frame at time $t$ is assigned 
the embedding of the caption segment whose temporal 
interval $[t^{(m)}_{j,\text{start}},\, 
t^{(m)}_{j,\text{end}})$ contains $t$, yielding a 
temporally aligned textual feature stream 
$x^{\text{text}}_t \in \mathbb{R}^{D}$. This 
alignment is a key design choice: rather than treating 
text as a global video descriptor, TASOT assigns 
fine-grained semantic context to individual frames, 
enabling the transport objective to exploit local 
semantic transitions as cues for segment boundaries.

\noindent\textbf{Modality-Specific Prototype Learning.}
TASOT learns $K$ normalized action prototypes 
$\mathbf{A} = \{a_k\}_{k=1}^{K} \subset 
\mathbb{R}^{D}$ that serve as soft cluster centroids 
within the optimal transport objective. Compared with 
ASOT~\cite{ASOT2024unsupervised}, which operates on a 
single visual feature stream through a shared projection 
head, TASOT introduces modality-specific projection heads 
for the visual and textual streams independently. 
Concretely, visual features $x^{\text{img}}_t$ and 
textual features $x^{\text{text}}_t$ are projected 
into a shared latent space through separate learnable 
linear projections followed by $\ell_2$ normalization, 
yielding $z^{\text{img}}_t$ and $z^{\text{text}}_t$ 
respectively:
\begin{equation}
    z^{\text{img}}_t = \frac{W_{\text{img}}\, 
    x^{\text{img}}_t}{\|W_{\text{img}}\, 
    x^{\text{img}}_t\|_2}, \quad
    z^{\text{text}}_t = \frac{W_{\text{text}}\, 
    x^{\text{text}}_t}{\|W_{\text{text}}\, 
    x^{\text{text}}_t\|_2}.
    \label{eq:projections}
\end{equation}
This design allows each modality to learn its own 
alignment to the shared prototype space, preventing 
the dominant visual modality from suppressing 
complementary semantic structure in the text stream, 
a failure mode observed when using feature 
concatenation, as confirmed in our ablation 
(Table~\ref{tab:modality_analysis}).

\noindent\textbf{Multimodal Transport Cost.}
Using the projected embeddings and shared prototypes, 
TASOT defines modality-specific cosine-distance cost 
matrices:
\begin{equation}
    C^{\text{img}}_{i,k} = 1 - \langle 
    z^{\text{img}}_i,\, a_k \rangle, \qquad
    C^{\text{text}}_{i,k} = 1 - \langle 
    z^{\text{text}}_i,\, a_k \rangle.
    \label{eq:unimodal_costs}
\end{equation}
These are fused at the \textit{cost level} into a 
single multimodal transport cost:
\begin{equation}
    C_{i,k} = \beta\, C^{\text{img}}_{i,k} + 
    (1-\beta)\, C^{\text{text}}_{i,k},
    \label{eq:mm_cost}
\end{equation}
where $\beta \in [0,1]$ controls the visual-text 
trade-off. Cost-level fusion is a deliberate design 
choice with an important geometric justification. 
TASOT builds on the Fused Unbalanced Gromov-Wasserstein 
(FUGW) optimal transport formulation~\cite{thual2022aligning}, 
which combines a Kantorovich matching cost, encoding 
visual similarity between frames and prototypes, with 
a Gromov-Wasserstein structural prior that enforces 
temporal consistency over the transport plan 
$\mathbf{T}^\star$, and an unbalanced penalty that 
relaxes the requirement for all action classes to be 
equally represented within a video. The FUGW solver 
operates directly on the cost matrix $\hat{\mathbf{C}}$ 
to compute $\mathbf{T}^\star$. By fusing modalities 
at the cost level, rather than in the feature space, 
TASOT ensures that both visual appearance and 
semantic textual cues jointly shape the matching cost 
fed into the FUGW objective, allowing the 
Gromov-Wasserstein temporal consistency prior to 
simultaneously regularize the alignment of both 
modalities. Feature concatenation, by contrast, 
conflates the two modalities before the transport 
geometry is defined, losing this structural advantage.

\noindent\textbf{Temporally Regularized Optimal 
Transport.}
A temporal prior $\mathbf{R}$ is incorporated into 
the cost matrix to encourage monotonic alignment 
between frame indices and action prototypes:
\begin{equation}
    \hat{C}_{i,k} = C_{i,k} + \rho\, R_{i,k}, 
    \qquad R_{i,k} = \left|\frac{i}{N} - 
    \frac{k}{K}\right|,
    \label{eq:temporal_prior}
\end{equation}
where $\rho \geq 0$ controls the strength of the 
temporal regularization. The resulting multimodal 
cost $\hat{\mathbf{C}}$ is passed to the FUGW OT 
solver, which minimizes the fused, unbalanced 
Gromov-Wasserstein objective:
\begin{equation}
\begin{aligned}
\mathbf{T}^\star
= \underset{\mathbf{T} \in \mathcal{T}_\mathbf{p}}{\arg\min}\;
&\alpha\, \mathcal{F}_{\text{GW}}(\mathbf{C}^v, \mathbf{C}^a, \mathbf{T}) \\
&+ (1-\alpha)\, \langle \hat{\mathbf{C}}, \mathbf{T} \rangle \\
&+ \lambda\, D_{\text{KL}}(\mathbf{T}^\top \mathbf{1}_N \| \mathbf{q})
\end{aligned}
\label{eq:fugw}
\end{equation}
where $\mathbf{C}^v$ and $\mathbf{C}^a$ encode 
temporal structure over frames and prototypes 
respectively, $\alpha$ balances the GW temporal 
consistency prior against the multimodal matching 
cost, and $\lambda$ controls the degree of unbalanced 
assignment, allowing a subset of action classes to 
be absent from a given video, which is essential for 
the long-tailed class distributions characteristic 
of surgical workflows.

The resulting soft transport plan 
$\mathbf{T}^\star \in \mathbb{R}^{N \times K}$ 
assigns each frame $i$ to action prototype $k$ with 
probability $T^\star_{ik}$, and serves as pseudo-labels 
for self-training. Frame-level cluster probabilities 
are computed from the fused latent embeddings and 
optimized via cross-entropy loss against the OT 
assignments, jointly updating both the projection 
heads and the action prototypes $\mathbf{A}$.

\section{Experimental Results and Discussion}
\subsection{Experimental Setup}

We evaluate TASOT on three public surgical datasets: 
Cholec80~\cite{cholec80-2017dataset} (80 videos, 
7 phases), AutoLaparo~\cite{AutoLapro2022dataset} 
(21 videos, 7 phases), and 
MultiBypass140~\cite{MultiByPass2024dataset} (140 
videos from Strasbourg and Bern, annotated with 12 
phases and 46 steps). For MultiBypass140, results 
are reported separately for each center following 
the official zero-shot evaluation protocol. 
Predicted clusters are aligned with ground-truth 
labels using Hungarian 
matching~\cite{ASOT2024unsupervised}. We report 
segmental F1 score~\cite{temporal2024survey} as 
the primary metric, consistent with prior work. 
In the primary experiments, the number of clusters 
$K$ is set equal to the total number of annotated 
classes per dataset, consistent with standard 
practice in OT-based unsupervised 
segmentation~\cite{ASOT2024unsupervised}.

\subsection{State-of-the-Art Comparisons}

\begin{table*}[!tb]
\caption{Phase and step recognition results on Cholec80~\cite{cholec80-2017dataset}, AutoLaparo~\cite{AutoLapro2022dataset}, and MultiBypass140~\cite{MultiByPass2024dataset}. TASOT is compared with zero-shot surgical video-language models and the annotation-free ASOT baseline. Best results are shown in \textbf{bold}. Step-level F1 is not reported for zero-shot methods because their pretraining protocols do not use step-level supervision.}
\centering
\small
\renewcommand{\arraystretch}{1.1}
\begin{tabular*}{\textwidth}{@{\extracolsep{\fill}}lcccccc}
\toprule
\multirow{2}{*}{Method}
& Cholec80
& AutoLaparo
& \multicolumn{2}{c}{BernByPass70}
& \multicolumn{2}{c}{StrasByPass70} \\
\cmidrule(lr){2-2}
\cmidrule(lr){3-3}
\cmidrule(lr){4-5}
\cmidrule(lr){6-7}
& Phase F1 & Phase F1
& Phase F1 & Step F1
& Phase F1 & Step F1 \\
\midrule
\rowcolor{gray!12}
\multicolumn{7}{c}{\textit{Zero-shot Methods}} \\
MIL-NCE~\cite{MIL-NCE2020end}        
    & 7.3  & 7.9  & 2.1  & \textemdash 
    & 3.1  & \textemdash \\
CLIP-SVL~\cite{CLIP2021ZeroShot}     
    & 19.6 & 16.2 & 7.1  & \textemdash 
    & 8.6  & \textemdash \\
SurgVLP~\cite{SurgVLP2025ZeroShot}   
    & 24.4 & 16.6 & 7.2  & \textemdash 
    & 6.9  & \textemdash \\
HecVL~\cite{HecVL2024ZeroShot}       
    & 26.3 & 18.9 & 13.6 & \textemdash 
    & 18.3 & \textemdash \\
PeskaVLP~\cite{PeskaVLP2024ZeroShot} 
    & 34.2 & 23.6 & 22.6 & \textemdash 
    & 28.6 & \textemdash \\
\rowcolor{gray!12}
\multicolumn{7}{c}{\textit{Annotation-free Methods}} \\
ASOT~\cite{ASOT2024unsupervised}     
    & 44.94  & 42.63  & 26.98  & 13.45
    & \textbf{56.11}  & 25.37 \\
TASOT (Ours)                         
    & \textbf{53.11} & \textbf{57.30}
    & \textbf{27.1} & \textbf{23.0} 
    & 52.3 & \textbf{30.7} \\
\bottomrule
\end{tabular*}
\label{tab:main_results}
\end{table*}

Table~\ref{tab:main_results} compares TASOT with recent zero-shot surgical video-language models and the annotation-free ASOT baseline. TASOT outperforms the strongest zero-shot baseline, PeskaVLP, across all phase recognition benchmarks, with gains of +18.9 F1 on Cholec80, +33.7 on AutoLaparo, +4.5 on BernByPass70, and +23.7 on StrasByPass70, demonstrating robustness across diverse surgical environments and centers. Relative to ASOT, TASOT improves phase recognition on Cholec80, AutoLaparo, and BernByPass70, and delivers clear gains in step recognition on MultiBypass140. In particular, TASOT improves step F1 from 13.45 to 23.0 on BernByPass70 and from 25.37 to 30.7 on StrasByPass70. Although ASOT achieves slightly higher phase F1 on StrasByPass70, the overall results show that TASOT’s multimodal formulation, which integrates textual cues into the OT framework, is particularly beneficial for more challenging settings and fine-grained step segmentation.



\subsection{Ablation Study}

\begin{table*}[!t]
\caption{Component ablation of TASOT on MultiBypass140. \textit{Video-only} uses DINOv3 features within ASOT, \textit{Text-only} uses aligned caption features only, and \textit{Feature concatenation} replaces cost-level fusion with feature-space fusion. \textit{TASOT (multimodal cost)} denotes the full model. Best results are shown in \textbf{bold}.}
\label{tab:modality_analysis}
\centering
\small
\setlength{\tabcolsep}{4pt}
\renewcommand{\arraystretch}{1.1}
\begin{tabularx}{\textwidth}{l *{4}{>{\centering\arraybackslash}X}}
\toprule
\multirow{2}{*}{Model}
& \multicolumn{2}{c}{BernByPass70}
& \multicolumn{2}{c}{StrasByPass70} \\
\cmidrule(lr){2-3}
\cmidrule(lr){4-5}
& Phase F1 & Step F1
& Phase F1 & Step F1 \\
\midrule
Video-only (DINOv3 + ASOT)      
    & \textbf{27.9} & 20.3  & 51.3  & 27.9 \\
Text-only                 
    & 18.2          & 7.98  & 26.3  & 9.39 \\
Feature concatenation     
    & 27.4          & 14.3  & 50.4  & 21.2 \\
\rowcolor{gray!12}
TASOT (multimodal cost)             
    & 27.1 & \textbf{23.0} & \textbf{52.3} & \textbf{30.7} \\
\bottomrule
\end{tabularx}
\end{table*}

\begin{table*}[!t]
\caption{Sensitivity to the multimodal weight $\beta$ on MultiBypass140. $\beta=1.0$ is visual-only, $\beta=0.0$ is text-only, and $\beta=0.8$ is used as the default setting in the remaining experiments. Best results are shown in \textbf{bold}.}
\label{tab:beta_sensitivity}
\centering
\small
\setlength{\tabcolsep}{4pt}
\renewcommand{\arraystretch}{1.08}
\begin{tabularx}{\textwidth}{l *{4}{>{\centering\arraybackslash}X}}
\toprule
\multirow{2}{*}{$\beta$}
& \multicolumn{2}{c}{BernByPass70}
& \multicolumn{2}{c}{StrasByPass70} \\
\cmidrule(lr){2-3}
\cmidrule(lr){4-5}
& Phase F1 & Step F1 & Phase F1 & Step F1 \\
\midrule
1.0 (visual-only) & \textbf{27.9} & 20.3 & 51.3 & 27.9 \\
\rowcolor{gray!12}
0.8 & 27.1 & \textbf{23.0} & \textbf{52.3} & \textbf{30.7} \\
0.5 & 25.6 & 12.93 & 42.0 & 16.17 \\
0.0 (text-only) & 18.2 & 7.98 & 26.3 & 9.39 \\
\bottomrule
\end{tabularx}
\end{table*}

We conduct ablations on MultiBypass140 to: (i) isolate 
the contribution of each component of TASOT, (ii) 
analyze sensitivity to the multimodal weight $\beta$, 
(iii) evaluate the choice of text encoder, and (iv) 
analyze the sensitivity of TASOT to the number of 
clusters $K$.

\noindent\textbf{Component isolation.}
Table~\ref{tab:modality_analysis}, together with the ASOT baseline in Table~\ref{tab:main_results}, provides a structured decomposition of TASOT's performance gains. Specifically, ASOT in Table~\ref{tab:main_results} represents the original OT framework with its original visual backbone, while \textit{Video-only (DINOv3 + ASOT)} in Table~\ref{tab:modality_analysis} isolates the effect of replacing that backbone with DINOv3 while still using no textual input. Comparing these two baselines shows that a substantial portion of the gain, particularly for step recognition, is attributable to the stronger visual encoder. Comparing \textit{Video-only (DINOv3 + ASOT)} with the full TASOT model then isolates the contribution of the multimodal cost and text branch.

Comparing the latter two rows shows that the multimodal cost provides clear gains for step recognition, improving Step F1 from 20.3 to 23.0 on BernByPass70 and from 27.9 to 30.7 on StrasByPass70. Feature concatenation consistently underperforms the multimodal cost formulation, particularly for step recognition, validating the benefit of cost-level fusion. Text-only performance is substantially lower across all settings, confirming that visual features remain the primary signal and that the text branch acts as a structured semantic complement rather than a standalone modality.

\noindent\textbf{Sensitivity to multimodal weighting.}
Table~\ref{tab:beta_sensitivity} analyzes the effect of the multimodal weight $\beta$ in Eq.~\ref{eq:mm_cost}. The results show that the best performance is obtained with an intermediate but vision-dominant setting ($\beta=0.8$), which consistently improves step recognition over the visual-only baseline while maintaining strong phase recognition. In contrast, equal weighting ($\beta=0.5$) leads to a clear degradation, and the text-only setting ($\beta=0.0$) performs substantially worse across all metrics. These results support the interpretation that textual cues act as a complementary signal that improves temporal segmentation when integrated with strong visual features, rather than as a standalone modality. Accordingly, we use $\beta=0.8$ as the default setting in all remaining experiments.

\noindent\textbf{Text encoder choice.}
Table~\ref{tab:Backbone_comparison} compares CLIP 
against Gemma~\cite{gemma2024text} as the text 
encoder, following recent contrastive learning 
practice~\cite{chen2025vl}. DINOv3 combined with 
CLIP consistently achieves the highest performance 
across all metrics. We attribute this to CLIP's 
contrastive pretraining on image-text pairs, which 
produces embeddings that are inherently aligned 
with visual features, a property that directly 
benefits the joint OT cost formulation in 
Eq.~\ref{eq:mm_cost}. Combining Gemma and CLIP 
via concatenation does not improve over CLIP 
alone, further supporting the use of cost-level 
multimodal fusion over feature-space combination.

\begin{table*}[!t]
\caption{Text encoder analysis on MultiBypass140 using DINOv3 visual features within TASOT. We compare CLIP, Gemma, and their feature-space concatenation for the textual branch. $\,\Vert\,$ denotes feature concatenation. Best results are shown in \textbf{bold}.}
\label{tab:Backbone_comparison}
\centering
\small
\setlength{\tabcolsep}{4pt}
\renewcommand{\arraystretch}{1.08}
\begin{tabularx}{\textwidth}{l l *{4}{>{\centering\arraybackslash}X}}
\toprule
\multirow{2}{*}{Visual Encoder}
& \multirow{2}{*}{Text Encoder}
& \multicolumn{2}{c}{BernByPass70}
& \multicolumn{2}{c}{StrasByPass70} \\
\cmidrule(lr){3-4} \cmidrule(lr){5-6}
& & Phase F1 & Step F1 & Phase F1 & Step F1 \\
\midrule
\rowcolor{gray!12}
DINOv3 & CLIP               
    & \textbf{27.1} & \textbf{23.0} 
    & \textbf{52.3} & \textbf{30.7} \\
DINOv3 & Gemma              
    & 26.7 & 22.4 & 49.2 & 29.9 \\
DINOv3 & Gemma $\,\Vert\,$ CLIP       
    & 26.6 & 22.1 & 50.3 & 30.6 \\
\bottomrule
\end{tabularx}
\end{table*}

\begin{table*}[!t]
\caption{Effect of video-specific cluster counts on MultiBypass140. The $k$-specific setting is an oracle analysis that uses the number of active classes per video and is not available in practical deployment. Best annotation-free results are shown in \textbf{bold}.}
\label{tab:cluster_sup}
\centering
\small
\setlength{\tabcolsep}{4pt}
\renewcommand{\arraystretch}{1.08}
\begin{tabularx}{\textwidth}{l *{4}{>{\centering\arraybackslash}X}}
\toprule
\multirow{1}{*}{Method}
& \multicolumn{2}{c}{BernByPass70}
& \multicolumn{2}{c}{StrasByPass70} \\
\cmidrule(lr){2-3} \cmidrule(lr){4-5}
& Phase F1 & Step F1 & Phase F1 & Step F1 \\
\midrule
\rowcolor{gray!12}
\multicolumn{5}{c}{\textit{Annotation-free Methods}} \\
TASOT                
    & 27.1 & 23.0 & 52.3 & 30.7 \\
TASOT ($k$-specific, oracle)  
    & \textbf{49.7} & \textbf{48.8} 
    & \textbf{56.9} & \textbf{52.4} \\
\rowcolor{gray!12}
\multicolumn{5}{c}{\textit{Supervised Methods}} \\
TeCNO~\cite{MultiByPass2024dataset}     
    & 59.2 & 47.5 & 80.7 & 58.1 \\
MTMS-TCN~\cite{MultiByPass2024dataset}  
    & 62.4 & 47.9 & 79.8 & 57.3 \\
\bottomrule
\end{tabularx}
\end{table*}

\begin{figure*}[!t]
    \centering
    \includegraphics[width=1\linewidth]{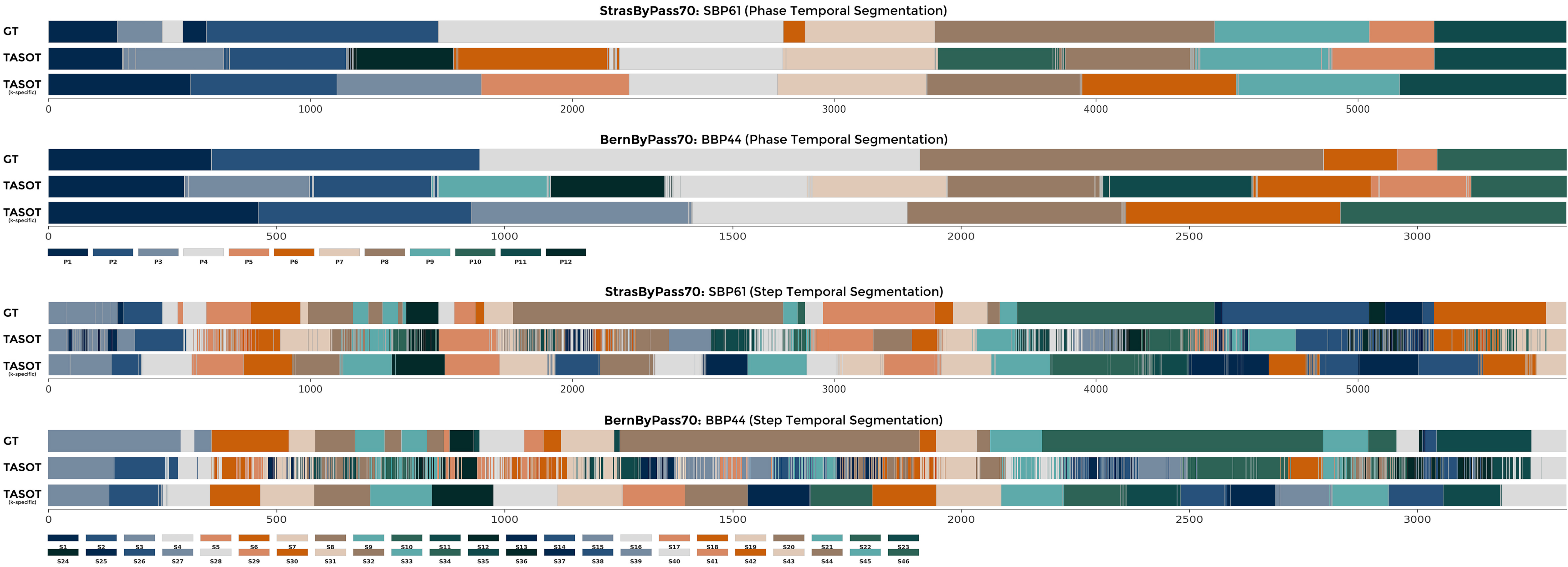}
    \caption{\textbf{Qualitative results.} 
    Ground-truth (GT) and TASOT predictions on 
    MultiByPass140. Phase-level segmentations 
    are temporally coherent with well-aligned 
    boundaries. Step-level segmentation is more 
    challenging due to finer temporal granularity 
    and a fixed cluster count $K$ that may 
    exceed the number of active steps in a 
    given video.}
    \label{fig:gt_plot}
\end{figure*}

\noindent\textbf{Sensitivity to cluster count $K$.}
In our primary experiments, $K$ is fixed to the 
total number of annotated classes per dataset. 
However, not all classes are necessarily present 
in every individual video, meaning the model may 
tend to predict more segments than actually 
occur. Table~\ref{tab:cluster_sup} reports 
performance when $K$ is adapted to the actual 
number of active classes per video 
($k$-specific). This is an \textit{oracle} 
analysis: it requires privileged knowledge of 
per-video class counts unavailable at inference 
time and does not represent a practical 
deployment scenario. It is reported solely to 
quantify the performance ceiling achievable with 
ideal cluster count estimation. Under this oracle 
setting, TASOT achieves substantial gains, with 
Step F1 increasing from 23.0 to 48.8 on 
BernByPass70 and from 30.7 to 52.4 on 
StrasByPass70, the latter nearly matching the 
supervised TeCNO baseline on steps. These results 
highlight adaptive cluster count estimation as a 
promising direction for future work, and confirm 
that the primary practical limitation of TASOT 
lies in $K$ selection rather than in the 
multimodal OT formulation itself.

\noindent\textbf{Discussion and limitations.}
While TASOT shows strong gains over recent zero-shot surgical video-language baselines and the ASOT baseline, these comparisons should be interpreted with appropriate context. In contrast to zero-shot video-language models, TASOT leverages video-conditioned textual descriptions generated at inference time, providing an additional source of semantic information that is specific to each input video. We therefore view the comparison as one between annotation-free multimodal temporal segmentation strategies rather than a strictly like-for-like architectural comparison. In addition, the gains from the text branch are most pronounced for fine-grained step recognition, while visual-only OT baselines remain competitive for some phase-level settings, such as StrasByPass70. Finally, TASOT still depends on the quality of generated captions and on the choice of a fixed cluster count $K$, both of which remain important directions for future improvement.

\section{Conclusion}

We presented TASOT, an annotation-free multimodal optimal 
transport framework for temporal segmentation in robotic 
surgical systems. By extending the ASOT formulation to 
incorporate temporally aligned textual descriptions 
generated directly from silent surgical video, TASOT 
introduces a principled cost-level fusion of visual and 
semantic cues within a unified Fused Unbalanced 
Gromov-Wasserstein objective---requiring no surgical-domain annotations, no manual task-specific labels, and no narrated footage. TASOT outperforms zero-shot 
surgical video-language baselines across four benchmark 
settings spanning laparoscopic and robotic procedures, 
demonstrating that fine-grained workflow understanding 
in diverse robotic surgical environments can be achieved 
without large-scale surgical-specific pretraining pipelines.

Ablation studies confirm three key findings. First, 
the multimodal text branch provides complementary 
gains beyond the stronger DINOv3 visual encoder, 
particularly for fine-grained step recognition, 
validating the role of textual cues in the multimodal 
cost formulation. Second, cost-level fusion within 
the OT objective outperforms feature concatenation 
on the more challenging step recognition tasks, 
supporting the geometric argument that fusing 
modalities before the transport geometry is defined 
can lose structural information important for 
temporally consistent segmentation. Third, the primary practical limitation 
of TASOT lies in fixed cluster count $K$: our oracle 
analysis shows that adapting $K$ to the actual number 
of active classes per video yields substantial gains, 
nearly matching supervised baselines on step recognition, identifying adaptive cluster count estimation as the most impactful direction for future work.

Beyond surgical robotics, TASOT's annotation-free 
multimodal OT formulation is directly applicable to 
any long, untrimmed procedural video domain where 
vision-language models can generate temporally grounded 
descriptions---including industrial assembly, 
skills training, and neurosurgical workflows. As 
vision-language models continue to improve in temporal 
grounding accuracy, the quality of the text branch 
is expected to improve correspondingly, further 
strengthening the practical case for annotation-free 
surgical AI in real robotic operating room deployments.


\end{document}